\documentclass[11pt]{article}
\usepackage{amsmath,amssymb,amsfonts,amsthm}
\usepackage{graphicx}
\usepackage[margin=0.8in]{geometry}
\usepackage{bm}
\usepackage{bbm}
\usepackage{mathtools}
\usepackage{xcolor}
\usepackage{algorithm}
\usepackage{algorithmic}
\usepackage{comment}
\usepackage[sort,comma]{natbib}
\usepackage{mathrsfs}
\usepackage{float}
\usepackage{booktabs}
\usepackage{array}
\usepackage{authblk}
\usepackage[colorlinks=true, linkcolor=blue, citecolor=blue, urlcolor=blue]{hyperref}
\usepackage{algorithm}
\usepackage{algorithmic}
\usepackage{comment}

\title{Non--negative matrix factorization using the \textit{R} package \textsf{nnmf}}

\author{Volkan Sevin\c{c}$^1$, Nikolas Kontemeniotis$^2$, Theodoros Perdikis$^3$ and Michail Tsagris$^4$ \\
$^1$ Mu\u{g}la S\i tk\i Ko\c{c}man University, Faculty of Science, Department of Statistics,  \\ K\"otekli Kamp\"us\"u, Mu\u{g}la, T\"urkiye \\
\href{mailto:vsevinc@mu.edu.tr}{vsevinc@mu.edu.tr} \\
$^2$ Department of Economics, University of Crete, Gallos Campus, Rethimno, Greece \\
\href{mailto:kontemeniotisn@gmail.com}{kontemeniotisn@gmail.com} \\ 
$^3$ Department of  Tourism Studies,University of Piraeus,Piraeus, Greece \\
\href{mailto:theoperdik@aueb.gr}{theoperdik@unipi.gr} \\
$^4$ Department of Economics \& Institute of Research in Education and Digital Social \\ Sciences and Humanities, University of Crete, Gallos Campus, Rethimno, Greece \\
\href{mailto:mtsagris@uoc.gr}{mtsagris@uoc.gr} }

\begin{document}
\maketitle

\begin{center}
\textbf{Abstract}
\end{center}
Non--negative matrix factorization (NMF) has become an established dimensionality reduction technique for extracting latent structures from non--negative data and has found widespread applications in fields such as bioinformatics, text mining, image analysis, and recommender systems. As the popularity of NMF has increased, numerous \textit{R} packages implementing different optimization strategies and computational frameworks have been developed. Despite their widespread availability, comprehensive evaluations of these implementations under real--world data conditions remain limited. Consequently, researchers often lack objective guidance when selecting an appropriate package for practical applications. This study introduces a new \textit{R} package for NMF and offers asystematic performance comparison with two widely available \textit{R} packages for NMF analysis. Rather than relying on simulated datasets, the evaluation is conducted using real--world data to better reflect the complexity, heterogeneity, and noise characteristics encountered in practical analytical settings. The packages are assessed using a consistent experimental framework, with emphasis on computational efficiency, convergence behavior, reconstruction accuracy, memory utilization, and the stability of the resulting matrix factorization.  \\
\\
\textbf{Keywords}: non--negative matrix factorization, lower--rank approximation, \textit{R} packages

\section{Introduction}
The continuous growth of high--dimensional data across diverse scientific and engineering disciplines has increased the demand for efficient analytical techniques capable of extracting meaningful information while reducing data complexity. Matrix factorization methods have become fundamental tools for this purpose because they transform complex datasets into lower--dimensional representations without substantially compromising their underlying structure. Among these methods, non--negative matrix factorization (NMF) has attracted considerable attention due to its ability to decompose a non--negative data matrix into lower--rank non--negative factors that often provide intuitive and interpretable representations of the original data. Unlike traditional dimensionality reduction approaches that permit both positive and negative values in their latent components, NMF imposes non--negativity constraints on both basis and coefficient matrices. This characteristic frequently produces parts--based representations that are more consistent with the inherent properties of many real--world datasets, thereby enhancing the interpretability of the extracted latent features.

The development of NMF has been shaped by a series of methodological advances aimed at improving the representation and analysis of non--negative high--dimensional data. Early foundations of this approach can be traced back to statistical factorization methods that incorporated non--negativity constraints and provided a basis for later NMF formulations. \citet{paatero1994} introduced positive matrix factorization (PMF), a statistical factor analysis framework that decomposes non--negative data matrices while accounting for measurement uncertainties. Building upon these concepts, \citet{lee1999} investigated non--negative matrix representations from a machine learning perspective and demonstrated that non--negativity constraints could generate meaningful parts--based representations. Subsequently, \citet{lee2000} proposed practical multiplicative update algorithms for NMF optimization, providing an efficient and widely applicable framework for matrix factorization problems. Their contributions played a fundamental role in establishing NMF as a widely adopted technique across diverse application domains.

\citet{chalise2017} developed integrative non--negative matrix factorization (intNMF) framework for the joint analysis of multiple high--dimensional omics datasets. The method enables molecular subtype identification by integrating information across different biological layers without requiring specific distributional assumptions, demonstrating its utility in cancer classification studies using TCGA data. \citet{quintero2020} developed two  novel \textit{R}--based tools that provide an intNMF workflow for genome--scale data analysis. The framework introduces \textit{TensorFlow}--based NMF solvers, a rational strategy for selecting the optimal factorization rank, and a feature selection approach to improve the extraction and biological interpretation of molecular signatures.

The increasing popularity of NMF has also been supported by computational environments such as \textit{R}, which has become one of the most widely used platforms for statistical computing and data analysis, offering an extensive ecosystem of packages for machine learning and matrix factorization. Consequently, various novel \textit{R} packages have been developed to implement NMF based on different computational frameworks and optimization strategies. Examples of these studies conducted within the last decade include the following.

\citet{lin2020} proposed a faster coordinate--wise descent algorithm for NMF that scales better to large datasets than existing methods. They also extended NMF to naturally handle missing values, introduced a new rank--selection method based on missing value imputation, and demonstrated how prior knowledge can be incorporated into NMF through masking for more meaningful decompositions. In addition to these methodological advances, they implemented all of their proposed algorithms in the \textit{R} package \textsf{NNLM}, making the methods publicly available\footnote{The package has been removed from CRAN's repository.}.

Furthermore, \citet{tsuyuzaki2023} introduced \textsf{nnTensor}, an \textit{R} package designed for non--negative matrix and tensor decomposition. The package provided a unified \textit{R} environment for non--negative decomposition methods, enabling researchers to apply advanced algorithms without relying on \textit{MATLAB}--based implementations.

More recently, \citet{carbonetto2026} established a formal connection between maximum--likelihood estimation in topic models and NMF, demonstrating that modern NMF optimization techniques can substantially improve the speed and quality of topic model fitting. They implement these methods in the \textit{R} package \textsf{fastTopics} \citep{carbonetto2023} that was developed specifically for count data, providing an efficient and scalable tool for analyzing large text datasets.

The \textsf{NMFN} \cite{liu2010} represents one of the earliest comprehensive implementations of NMF within the \textit{R} environment. The package provides a unified computational framework for decomposing non--negative matrices into lower--dimensional latent representations while incorporating several classical optimization algorithms that have been extensively used in the NMF literature. The \textsf{hNMF} \citep{gaujoux2015} extends the conventional NMF framework by incorporating hierarchical decomposition into the factorization process. The major drawback of both packages is their low computational efficiency. 

Two efficiently written \textit{R} packages are the \textsf{NMF} \citet{gaujoux2010} and the \textsf{RcppML} \citep{debruine2026}, both containing \text{C++} implementations of NMF. The \textsf{NMF} package constitutes one of the most comprehensive and widely adopted software frameworks for NMF in the \textit{R} statistical environment. The package was designed to provide a standardized, extensible, and reproducible platform that enables users to perform matrix factorization, compare alternative optimization algorithms, evaluate model quality, and reproduce computational experiments under consistent analytical settings. The \textsf{RcppML} package is a high--performance computational framework designed for NMF and related matrix decomposition techniques within the \textit{R} statistical computing environment. The package addresses the computational limitations commonly encountered in large--scale NMF analyses by integrating optimized \textit{C++} implementations through the \textsf{Rcpp} interface, thereby substantially improving computational efficiency while maintaining seamless compatibility with conventional \textit{R}--based analytical workflows \citep{eddelbuettel2011}. Owing to its efficient implementation, \textsf{RcppML} supports both dense and sparse matrix representations, making it particularly suitable for high--dimensional datasets encountered in fields such as bioinformatics, transcriptomics, text mining, and machine learning.

Another important limitation in the existing literature is that performance evaluations frequently rely on simulated datasets. While simulation studies provide controlled experimental conditions, they may not adequately capture the heterogeneity, noise, sparsity, and structural complexity characteristic of real--world data. Consequently, conclusions obtained from simulated experiments may not fully reflect practical performance under realistic analytical conditions. Benchmarking studies based on authentic datasets therefore provide a more informative basis for evaluating the practical strengths and limitations of competing implementations.

The present paper introduces a new \textit{R} package for NMF the \textsf{nnmf} package \citep{tsagris2026}. The implementation of the NMF is based on \textit{C++}, rendering it computationally efficient. The \textsf{nnmf} package provides a flexible and computationally efficient implementation of NMF in R. The package was designed to accommodate a wide range of matrix factorization applications involving dense, sparse, and high--dimensional datasets while offering multiple optimization strategies that improve algorithmic flexibility and computational performance.  

This paper further performs a systematic comparison of the \textsf{nnmf} package to \textsf{NMF} and \textsf{RcppML}. Unlike studies relying primarily on simulated data, the proposed evaluation is conducted using real--world datasets to better represent practical analytical environments. All packages are assessed under a unified experimental framework using identical datasets and comparable parameter settings to ensure a fair and reproducible comparison. Performance is evaluated with respect to computational time, memory utilization, convergence behavior, reconstruction accuracy, and factorization stability.

The primary objective of this study is to identify a universally superior implementation, both in terms of accuracy and computational cost. The findings are expected to provide practical guidance for researchers selecting NMF software in the \textit{R} environment while contributing to a more comprehensive understanding of how different algorithmic implementations perform in real--world applications. By establishing a reproducible benchmarking framework based on authentic datasets, this work also seeks to support future methodological developments and comparative studies in NMF research.

\section{Materials and Methods}
This section describes the methodological framework employed for matrix decomposition and latent feature extraction. The theoretical principles of matrix factorization are introduced first, followed by the fundamentals of NMF and the optimization algorithms developed for solving the NMF problem. These approaches were selected due to their ability to identify hidden structures in high--dimensional datasets while providing low--dimensional representations with improved interpretability.

\subsection{Matrix factorization}
Matrix factorization (MF) is a fundamental mathematical framework used for extracting latent structures from high--dimensional datasets by decomposing an original matrix into a product of lower--dimensional matrices. This approach assumes that an observed data matrix can be represented using a limited number of underlying factors that capture dominant patterns and relationships among variables. Due to its ability to reduce dimensional complexity while preserving meaningful information, MF has been extensively applied in various fields, including signal processing, bioinformatics, recommender systems, image analysis, text mining, and hyperspectral data interpretation.

Let $\bm{X} \in \mathbb{R}^{n \times p}$ denote a data matrix containing $n$ observations and $p$ variables. In conventional matrix factorization, the objective is to approximate $\bm{X}$ using two lower--rank matrices:
\begin{equation*}
\bm{X} \approx \bm{W H}
\end{equation*}
where $\bm{W} \in \mathbb{R}^{n \times k}$ and $\bm{H} \in \mathbb{R}^{k \times p}$ represent factor matrices, and $k$ corresponds to the selected latent dimensionality, generally satisfying $k \ll \min(n,p)$. The columns of $\bm{W}$ and $\bm{H}$ describe latent components and their corresponding contributions to the original data representation. The $\bm{W}$ is called basis or the lower--rank representation of the matrix $\bm{X}$, and $\bm{H}$ is the matrix loadings or coefficients.  

The quality of factorization is commonly evaluated by minimizing the reconstruction error between the original matrix and its approximation. A frequently used optimization criterion is based on the Frobenius norm:
\begin{equation} \label{frob}
\min_{\bm{W},\bm{H}} \; \lVert \bm{X} - \bm{W H}^\top \rVert_F^{2}
\end{equation}
where $\lVert \rVert_F$ represents the Frobenius norm. Depending on the characteristics of the dataset and the intended application, additional constraints or regularization terms may be incorporated into the optimization process to improve robustness, interpretability, and generalization capability.

Although conventional MF techniques are effective for dimensionality reduction, their interpretability can be limited because the factor matrices may contain both positive and negative values. This characteristic can complicate the physical interpretation of extracted components, particularly in applications where measured variables represent additive contributions or concentration--related phenomena. To overcome this limitation, constrained factorization approaches, particularly NMF, have been developed.

\subsection{Non--negative matrix factorization}
NMF is an extension of MF that introduces non--negativity constraints on the decomposed matrices. Unlike conventional MF, NMF restricts all elements of the factor matrices to non--negative values, thereby generating additive and physically interpretable representations of the original dataset.

For a non--negative data matrix $\bm{X} \in \mathbb{R}_{+}^{n \times p}$, NMF aims to identify two non--negative matrices $\bm{W}$ and $\bm{H}$ such that:
\begin{equation*}
\bm{X} \approx \bm{W H}
\end{equation*}
where:
\begin{equation*}
\bm{W} \geq \bm{0}, \quad \bm{H} \geq \bm{0}
\end{equation*}

The matrix $\bm{W}$ contains basis vectors representing characteristic patterns or components, whereas $\bm{H}$ describes the contribution of these components for each observation. Because the decomposition is based on additive combinations of non--negative factors, NMF is particularly advantageous for datasets in which negative contributions have no meaningful physical interpretation.

The optimization problem in NMF is generally formulated as minimizing the difference between the original matrix and its reconstructed approximation. The most commonly employed objective functions include Euclidean distance and Kullback--Leibler divergence. The Euclidean formulation can be expressed as:
\begin{equation} \label{sse}
\min_{\bm{W},\bm{H}} \; \tfrac{1}{2} \lVert \bm{X} - \bm{W H} \rVert_F^{2} \ \ \text{
subject to} \ \ W_{ij} \geq 0, \ H_{ij} \geq 0, \ \forall \ i,j.
\end{equation}

The imposed non--negativity constraint provides several advantages, including enhanced interpretability of latent components, sparse representation of complex datasets, and improved identification of meaningful patterns. Consequently, NMF has become a widely adopted technique in areas such as molecular profiling, metabolomics, chemical spectroscopy, remote sensing, and machine learning.

However, NMF optimization is inherently challenging because the objective function is non--convex with respect to $\bm{W}$ and $\bm{H}$ simultaneously. Therefore, iterative optimization algorithms are required to estimate the factor matrices, and different algorithmic strategies have been proposed to improve convergence speed, reconstruction accuracy, and stability.

\subsection{NMF related \textit{R} packages}
The packages were selected based on their methodological diversity, computational characteristics, implementation strategies, and widespread adoption within the \textit{R} ecosystem. Although all packages implement NMF, they differ substantially in terms of optimization algorithms, computational efficiency, scalability, support for sparse matrices, parallel computing capabilities, and additional analytical functionalities. A concise overview of the main characteristics and methodological features of each package is provided in the following sections.

These \textit{R} packages represent diverse computational and methodological strategies for addressing different analytical requirements. The \textsf{NMF} package provides a comprehensive environment that integrates multiple NMF algorithms with model evaluation, visualization, and analytical utilities. In comparison, \textsf{RcppML} focuses on improving computational performance and scalability through optimized \textit{C++} implementations, making NMF more applicable to large--scale and high--dimensional datasets. The \textsf{nnmf} package introduces additional optimization strategies, providing users with greater flexibility in selecting appropriate algorithms for different data characteristics and analytical objectives. Together, these packages demonstrate the evolution of NMF implementations in \textit{R} from basic algorithmic applications toward more scalable, flexible, and specialized computational frameworks.

Although these packages have been widely adopted, selecting an appropriate implementation remains a practical challenge. Their algorithmic diversity leads to differences in computational efficiency, convergence behavior, memory requirements, numerical stability, and reconstruction quality. Furthermore, package documentation generally emphasizes methodological aspects rather than providing comprehensive empirical comparisons across implementations. Consequently, researchers often select a package based on familiarity or availability instead of objective performance evidence.

\subsubsection{The \textsf{NMF} package}
Its modular architecture supports a broad collection of optimization algorithms, including multiplicative update rules, alternating least squares approaches, and numerous algorithmic variants proposed in the literature. By implementing these methods through a unified interface, the package enables objective comparisons among optimization algorithms while minimizing implementation--related variability \citep{gaujoux2010}. In addition, the framework allows users to incorporate custom optimization routines, making the package particularly valuable for methodological research and algorithm development.

One of the principal strengths of the NMF package is its extensive framework for determining the optimal factorization rank. Because the selection of matrix rank directly influences clustering performance, latent structure identification, and biological interpretation, the package provides multiple complementary evaluation criteria. These include the cophenetic correlation coefficient, sum of squares of errors (SSE) (or squared Frobenius norm), dispersion statistics, silhouette width, and consensus clustering measures \citep{brunet2004}. The combined use of these quantitative metrics improves the robustness of model selection and facilitates more reliable interpretation of factorization results.

The package additionally emphasizes computational reproducibility through standardized random seed management, comprehensive recording of algorithmic parameters, and support for parallel computing. These features enable exact replication of computational analyses while substantially reducing execution time for repeated factorization procedures, which are frequently required when evaluating solution stability in high--dimensional biological datasets \citep{gaujoux2010}.

\subsubsection{The \textsf{RcppML} package}
The computational framework implemented in \textsf{RcppML} is centered on efficient optimization of the standard NMF model. Unlike conventional \textit{R} implementations that primarily rely on interpreted code, \textsf{RcppML} performs the computationally intensive optimization procedures using compiled \textit{C++} routines. This design substantially reduces execution time during iterative optimization while improving scalability for large datasets. Such computational efficiency is particularly advantageous in applications requiring repeated matrix factorization, including parameter tuning, stability assessment, and bootstrap--based validation.

Beyond matrix factorization, the package incorporates additional machine learning utilities, including divisive clustering and optimized matrix operations that facilitate downstream statistical analyses \citep{debruine2026}. Furthermore, its compatibility with the broader \textit{R} ecosystem enables straightforward integration with preprocessing, visualization, clustering, and predictive modeling workflows. This interoperability enhances reproducibility by allowing complete analytical pipelines to be implemented within a unified computational environment.

\section{The NMF implementation in the \textsf{nnmf} package}
A distinguishing characteristic of \textsf{nnmf} is its support for several optimization algorithms, including approaches based on alternating non--negativity--constrained least squares and related numerical optimization techniques. The availability of multiple optimization strategies enables users to select algorithms according to the structural characteristics of the dataset and the desired convergence behavior. Since convergence rate and numerical stability represent critical factors affecting NMF performance, algorithmic flexibility provides an important advantage when analyzing complex datasets \citep{kim2008, wang2013}. The package also incorporates parallel computing capabilities that significantly reduce computational time during iterative optimization. arallelization is particularly beneficial when repeated factorizations are required for model selection, stability analyses, or cross--validation procedures involving large--scale matrices.

Within the \textit{R} ecosystem, \textsf{nnmf} integrates naturally with established analytical workflows, allowing matrix factorization to be combined with data preprocessing, visualization, clustering, and subsequent statistical modeling. Consequently, the package is well suited for applications in computational biology, bioinformatics, machine learning, and data mining, where NMF is widely employed for feature extraction, latent pattern identification, and dimensionality reduction. Support for sparse matrix representations further enhances its applicability to modern datasets characterized by increasing dimensionality and computational complexity.

\subsection{The QP and FEG algorithms}
There are several algorithms to perform NMF minimizing the Frobenius norm. The one adopted by \textsf{nnmf} is the alternating nonnegative least squares (ANLS) \citep{kim2008nonnegative, kim2011algorithms}, where each matrix is updated in an alternating fashion:
\begin{align*}
\bm{H}^{(t+1)} &= \arg\min_{\bm{H} \geq 0} \|\bm{X} - \bm{W}^{(t)} \bm{H}\|_F^2 \\
\bm{W}^{(t+1)} &= \arg\min_{\bm{W} \geq 0} \|\bm{X}^\top - \bm{H}^{(t+1)\top} \bm{W}^\top\|_F^2.
\end{align*}

One matrix is fixed and the other is updated, and this process is iterated until convergence. There are a few packages that perform NMF, with the most popular one being the \textsf{NMF} \citep{gaujoux2010}. We have the option to perform either quadratic programming (QP) or the exponentiated gradient (EG) algorithm. When QP is implemented, the non--negative least squares problem is solved using the algorithm of \cite{bro1997}.

\subsubsection{The QP algorithm: the $n > p$ case}
For the case of $n>p$, QP is solely utilized, and the steps of the ANLS are described below.
\paragraph{Updating $\bm{W}$ (row--wise QP):}
Fixing $\bm{H}$, let $\bm{x}_i^\top$ denote the $i$--th row of $\bm{X}$. Updating the $i$--th row
$\bm{w}_i \in \mathbb{R}^k$ of $\bm{W}$ solves
\begin{equation*}
\min_{\bm{w}_i \geq 0} \; \|\bm{x}_i^\top - \bm{w}_i^\top \bm{H}\|^2
= \min_{\bm{w}_i \geq 0} \; \bm{w}_i^\top (\bm{HH}^\top) \bm{w}_i - 2 \bm{x}_i^\top \bm{H}^\top \bm{w}_i + \text{const}.
\end{equation*}
This is a QP of the form
\begin{equation*}
\text{minimize} \quad \tfrac{1}{2} \bm{w}_i^\top \bm{G}_W \bm{w}_i + \bm{g}_i^\top \bm{w}_i
\qquad \text{subject to} \quad w_{i\ell} \geq 0, \; \ell = 1,\dots,k,
\end{equation*}
with
\begin{equation*}
\bm{G}_W = 2\bm{HH}^\top + \lambda \bm{I}_k, \qquad \bm{g}_i = -2\bm{Hx}_i,
\end{equation*}
where $\bm{I}_k$ is the $k$--dimensional identity matrix, and the ridge term $\lambda \bm{I}_k$ ($\lambda \approx 10^{-8}$) guards against rank--deficiency of $\bm{HH}^\top$. This is solved independently for each $i = 1,\dots,n$.

\paragraph{Updating $\bm{H}$ (single joint QP):}
Fixing $\bm{W}$, the objective is
\begin{equation*}
\|\bm{X} - \bm{WH}\|_F^2 = \|\bm{X}\|_F^2 - 2\,\mathrm{tr}(\bm{X}^\top \bm{WH}) + \mathrm{tr}(\bm{H}^\top \bm{W}^\top \bm{WH}).
\end{equation*}
Writing $\bm{h} = \mathrm{vec}(\bm{H}) \in \mathbb{R}^{kp}$ (columns of $\bm{H}$ stacked), and using
$\mathrm{tr}(\bm{H}^\top \bm{W}^\top \bm{WH}) = \bm{h}^\top \left(\bm{I}_p \otimes \bm{W}^\top \bm{W}\right) \bm{h}$ and $\mathrm{tr}(\bm{X}^\top \bm{WH}) = \bm{h}^\top \bm{d}$ with $\bm{d} = \mathrm{vec}(\bm{W}^\top \bm{X})$, the problem
becomes the single QP
\begin{equation*}
\text{minimize} \quad \tfrac{1}{2} \bm{h}^\top \bm{\Omega h} - \bm{d}^\top \bm{h}
\qquad \text{subject to} \quad \bm{h} \geq \bm{0},
\end{equation*}
with
\begin{equation*}
\bm{\Omega} = \bm{I}_p \otimes (\bm{W}^\top \bm{W}) \in \mathbb{R}^{kp \times kp}, \qquad
\bm{d} = \mathrm{vec}(\bm{W}^\top \bm{X}).
\end{equation*}
If $\bm{\Omega}$ is numerically singular, it is replaced by its nearest positive--definite approximation, $\mathrm{nearPD}(\bm{\Omega})$, before solving. Letting $\bm{h}^\ast$ denote the solution and $f^\ast = \tfrac{1}{2}\bm{h}^{\ast\top}\bm{\Omega h}^\ast - \bm{d}^\top \bm{h}^\ast$ the optimal QP value, the reconstruction error is recovered without recomputing $\|\bm{X}-\bm{WH}\|_F^2$
directly:
\begin{equation*}
\|\bm{X} - \bm{WH}\|_F^2 \;=\; \|\bm{X}\|_F^2 + 2f^\ast.
\end{equation*}
This joint formulation is only feasible because dropping it lets the $p$ separate row--blocks of $\bm{H}$ collapse into one QP with a block--diagonal (Kronecker) Hessian, rather than $p$ independent QPs.

\subsubsection{Hybrid QP--EG algorithm: the $n < p$ case}
When $n<p$ case, the joint QP for $\bm{H}$ above involves a $kp \times kp$ Hessian, which becomes computationally prohibitive. The $\bm{W}$--update is unchanged, but $\bm{H}$ is instead updated via EG.

\paragraph{Updating $\bm{H}$ (EG):}
The gradient of the Frobenius objective with respect to $H$ is
\begin{equation*}
\nabla_{\bm{H}} \|\bm{X} - \bm{WH}\|_F^2 = 2\bm{W}^\top(\bm{WH} - \bm{X}) = 2\bm{W}^\top \bm{E}, \qquad \bm{E} = \bm{WH} - \bm{X},
\end{equation*}
and the multiplicative update is
\begin{equation*}
\bm{H} \leftarrow \bm{H} \odot \exp\left(-\eta_H\, \bm{g}_H\right), \qquad \bm{g}_H = \bm{W}^\top \bm{E},
\end{equation*}
where $\odot$ denotes element--wise multiplication and $\eta_H > 0$ is the learning rate. The exponential map alone is sufficient to keep $\bm{H}$ element-wise non--negative. The reconstruction error SSE and the algorithm again terminates when $\big|SSE^{(t-1)} - SSE^{(t)}\big| < \tau$.

\begin{algorithm}
\caption{FQP algorithm for $n > p$}
\label{qp}
\begin{algorithmic}[1]
\REQUIRE Data matrix $\bm{X} \in \mathbb{R}^{n \times p}$, rank $k < p$, max iterations $T$, tolerance $\tau$, ridge parameter $\lambda > 0$
\ENSURE $n \times k$ matrix $\bm{W} \geq \bm{0}$, $k \times p$ matrix $\bm{H} \geq \bm{0}$
\STATE \textbf{Initialize:}
\STATE $\bm{W}^{(0)}, \bm{H}^{(0)} \gets$ non--negative initialization (K--means--based or uniform random)
\STATE $\bm{A}_W \gets \bm{I}_k$, $\; \bm{b}_W \gets \bm{0}_k$ \textcolor{gray}{// constraint matrix/vector for $\bm{W}$: $\bm{w}_i \geq \bm{0}$}
\STATE $\bm{A}_H \gets \bm{I}_{kp}$, $\; \bm{b}_H \gets \bm{0}_{kp}$ \textcolor{gray}{// constraint matrix/vector for $\mathrm{vec}(\bm{H})$}
\STATE Precompute $\|\bm{X}\|_F^2$
\FOR{$t = 1$ to $T$}
    \STATE \textbf{Update $W$:} \textcolor{gray}{//each row solved independently}
    \STATE Compute $\bm{G}_W = 2\bm{HH}^\top + \lambda \bm{I}_k$
    \STATE Compute $\bm{g}_W = 2\bm{HX}^\top \in \mathbb{R}^{k \times n}$
    \FOR{$i = 1$ to $n$}
        \STATE Solve QP: $\bm{w}_i^\ast = \arg\min_{\bm{w}} \tfrac{1}{2}\bm{w}^\top \bm{G}_W \bm{w} - \bm{g}_{W}[:,i]^\top \bm{w}$ \; subject to $\bm{A}_W^\top \bm{w} \geq \bm{b}_W$
        \STATE $\bm{W}_{i,:} \gets |\bm{w}_i^\ast|$
    \ENDFOR
    \STATE \textbf{Update $H$:} \textcolor{gray}{// single joint QP over $\mathrm{vec}(H) \in \mathbb{R}^{kD}$}
    \STATE Compute $d = \mathrm{vec}(\bm{W}^\top \bm{X})$
    \STATE Compute $\bm{\Omega} = \bm{I}_p \otimes (\bm{W}^\top \bm{W})$ \textcolor{gray}{// $kp \times kp$ Hessian, Kronecker form}
    \STATE Solve QP: $\bm{h}^\ast = \arg\min_{\bm{h}} \tfrac{1}{2}\bm{h}^\top \bm{\Omega h} - \bm{d}^\top \bm{h}$ \; subject to $\bm{A}_H^\top \bm{h} \geq \bm{b}_H$
    \STATE \quad (if $\bm{\Omega}$ is numerically singular, substitute $\mathrm{nearPD}(\bm{\Omega})$)
    \STATE $\bm{H} \gets |\mathrm{reshape}(\bm{h}^\ast, k, p)|$
    \STATE Compute objective: $SSE = \|\bm{X}\|_F^2 + 2 f^\ast$ \textcolor{gray}{// $f^\ast$ = optimal QP value from $H$--step}
    \IF{$|SSE^{(t-1)} - SSE^{(t)}| < \tau$}
        \STATE \textbf{break}
    \ENDIF
\ENDFOR
\STATE \textbf{return} $\bm{W}, \bm{H}, \bm{Z} = \bm{WH}, SSE, t$
\end{algorithmic}
\end{algorithm}

\begin{algorithm}
\caption{Hybrid QP--EG algorithm for $n < p$}
\label{qpeg}
\begin{algorithmic}[1]
\REQUIRE Data matrix $\bm{X} \in \mathbb{R}^{n \times p}$, rank $k < p$, learning rate $\eta_H > 0$, max iterations $T$, tolerance $\tau$, ridge parameter $\lambda > 0$
\ENSURE $n \times k$ matrix $\bm{W} \geq \bm{0}$, $k \times p$ matrix $\bm{H} \geq \bm{0}$
\STATE \textbf{Initialize:}
\STATE $\bm{W}^{(0)}, \bm{H}^{(0)} \gets$ non--negative initialization (K--means--based or uniform random)
\STATE $\bm{A}_W \gets \bm{I}_k$, $\; \bm{b}_W \gets \bm{0}_k$
\FOR{$t = 1$ to $T$}
    \STATE \textbf{Update $\bm{W}$:} \textcolor{gray}{// identical QP step as Algorithm 1}
    \STATE Compute $\bm{G}_W = 2\bm{HH}^\top + \lambda \bm{I}_k$
    \STATE Compute $\bm{g}_W = 2\bm{HX}^\top \in \mathbb{R}^{k \times n}$
    \FOR{$i = 1$ to $n$}
        \STATE Solve QP: $\bm{w}_i^\ast = \arg\min_{\bm{w}} \tfrac{1}{2}\bm{w}^\top \bm{G}_W \bm{w} - \bm{g}_{W}[:,i]^\top \bm{w}$ \; subject to $\bm{A}_W^\top \bm{w} \geq \bm{b}_W$
        \STATE $\bm{W}_{i,:} \gets |\bm{w}_i^\ast|$
    \ENDFOR
    \STATE \textbf{Update $\bm{H}$ via EG:} 
    \STATE Compute reconstruction: $\bm{Z} = \bm{WH}$
    \STATE Compute residual: $\bm{E} = \bm{Z} - \bm{X}$
    \STATE Compute gradient: $\bm{g}_H = \bm{W}^\top \bm{E} \in \mathbb{R}^{k \times D}$
    \STATE $\bm{H} \gets \bm{H} \odot \exp(-\eta_H \bm{g}_H)$ \textcolor{gray}{// element--wise; non--negativity automatic}
    \STATE Recompute $\bm{Z} = \bm{WH}$
    \STATE Compute objective: $SSE = \|\bm{X} - \bm{Z}\|_F^2$
    \IF{$|SSE^{(t-1)} - SSE^{(t)}| < \tau$}
        \STATE \textbf{break}
    \ENDIF
\ENDFOR
\STATE \textbf{return} $\bm{W}, \bm{H}, \bm{Z}, SSE, t$
\end{algorithmic}
\end{algorithm}

\subsection{Starting values}
One approach is to generate random matrices, from the uniform distribution. Another approach to initiate the $\bm{W}$ and $\bm{H}$ matrices is by using the $K$--means algorithm employing the algorithm of Hartigan--Wong \citep{hartigan1979}, without restarts and with 10 maximum iterations allowed\footnote{These are the default values for \textit{R}'s built-in function \texttt{kmeans()}.}. With large sample sizes (of the order of millions or more), the mini--batch $K$--means is another, faster, alternative. Another option is to perform the classical NMF and then normalize the produced matrices. When $n<p$, the standard $K$--means algorithm could be used to initialize the matrix $\bm{H}$, but preference is given to the high--dimensional sparse $K$--means algorithm of \cite{witten2010}. It must be highted that there is no unique solution in this $n<p$ case scenario.

\subsection{Prediction of new values}
Prediction of new values is straightforward, yet it requires QP. Without covariates, the matrix of loadings $\bm{H}$ are used to estimate the scores $\bm{W}^*$ of a set of new values $\bm{X}^*$ 
\begin{equation*}
\min_{\bm{W}^*} \left\| \bm{X}^* -\bm{W}^* \bm{H} \right\|_F^2.
\end{equation*}
The matrix of loadings $\bm{H}$ is already estimated and is not re--estimated. The same applies for the matrix of regression coefficients $\bm{B}$. For either case, the minimization is performed for each $\bm{w}^*_i$, $i=1,\ldots,n^*$, where $n^*$ denotes the sample size of the new values. 

\subsection{The commands in \textsf{nnmf}}
The package contains eight functions related to NMF, presented in Table \ref{tab:commands}, including the option to perform NMF with sparse and large--scale data, NMF minimizing the Manhattan distance, cross--validation and NMF with covariates. 

The function \texttt{init()} is used to initialize the $\bm{H}$ matrix via $K$--means, the \texttt{nmf.hals()} performs hierarchical ANLS, where instead of each row of $\bm{W}$ each column of $\bm{W}$ is estimated. The function is fast, but the results are highly variable. The \texttt{nmf.manh()} minimizes the Manhattan distance instead of the Euclidean distance. The function \texttt{nmf.sqp} minimizes the Frobenius norm and it is used for large--scale sparse data. The QP is solved using the \textit{R} package \textsf{osqp} \citep{osqp2024} and accepts a \textit{dgC} class sparse matrix. The \texttt{nmfqp.cv()} performs cross--validation to select the optimal value of $k$, while the function \texttt{nmfqp.pred} predicts the $\bm{W}$ for new values and the function \texttt{nmfqp.reg} performs NMF with covariates. 

The basic function, that will be used in the experimental design, is the \texttt{nmf.qp()} that  implements the Frobenius--norm based NMF. When the number of variables $p$ exceeds the sample size $n$, the hybrid scheme is used (Algorithm \ref{qpeg}), otherwise the plain QP (Algorithm \ref{qp}) is used. 
\begin{verbatim}
nmf.qp(X, k, H_init = NULL, k_means = TRUE, bs = 1, lr_h = 0.1, tol = 1e-6, 
maxiter = 1000, ridge = 1e-8, ncores = 1)
\end{verbatim}

\begin{table}[h!]
\centering
\small
\begin{tabular}{@{}ll@{}}
\toprule
\textbf{Command} & \textbf{Description} \\
\midrule
\textit{init()} & $K$--means--based initialization of the $\bm{H}$ matrix. \\
\textit{nmf.hals()} & NMF (Frobenius norm) via hierarchical ANLS. \\
\textit{nmf.manh()} & NMF minimizing the Manhattan distance; robust to outliers. \\
\textit{nmf.qp()} & NMF (Frobenius norm) via QP when $n>p$ and a hybrid of QP and EG when $n<p$. \\
\textit{nmf.sqp()} & NMF (Frobenius norm) via sequential QP for large sparse matrices. \\
\textit{nmfqp.cv()} & $K$--fold cross--validation for choosing the optimal rank $k$. \\
\textit{nmfqp.pred()} & Prediction of $\bm{W}$ and $\bm{Z}$ for new observations from a fitted $\bm{H}$. \\
\textit{nmfqp.reg()} & NMF with covariates $\bm{S}$: $\bm{X} \approx \bm{SB} + \bm{WH}$, fit by QP. \\
\bottomrule
\end{tabular}
\caption{The commands of the \textsf{nnmf} package.}
\label{tab:commands}
\end{table}

\begin{itemize}
  \item[X:] An $n \times p$ numerical matrix of data with non--negative entries.
  \item[k:] The rank of the factorization, $k \le D - 1$.
  \item[H\_init:] Optional initial value for $\bm{H}$. 
  \item[k\_means:] If \textit{TRUE}, it employs the $K$--means to initialize the matrix $\bm{H}$.
  \item[bs:] Batch size for mini--batch $K$--means initialization. This is suitable when the sample size is at the order of tens of millions or higher. 
  \item[lr\_h:] Learning rate for the EG algorithm update of $\bm{H}$ when $n<p$.
  \item[tol:] Tolerance value for convergence.
  \item[maxiter:] Maximum number of iterations.
  \item[ridge:] Small ridge constant added to the diagonal of the QP $\bm{I}_k$ matrix for numerical stability.
  \item[ncores:] Number of cores for the parallel update of $\bm{W}$. This is suitable when the sample size is large.
\end{itemize}

The function returns a list with the estimated $\bm{W}$ and $\bm{H}$ matrices, their product, the SSE (squared Frobenius norm), the number of iterations performed and the running time (in seconds). 

\section{Experimental design}
In this Section, the use of the methods presented above is illustrated by considering several datasets (presented in Table \ref{tab:datasets_info1}) obtained from  the \href{https://www.kaggle.com}{Kaggle platform} and from the \href{https://archive.ics.uci.edu}{University of California, Irvine Machine Learning Repository}. 
It should be highlighted that, for all the models being investigated, running the same dataset multiple times will yield different results. In order to find the best model that fits the data we propose to apply each model $B$ times and choose the one that gives the minimum SSE. Regarding the choice of $B$, note that, besides the efficiency of a model with respect to fitting well in a dataset, it is also important to take into consideration the computational cost. As a consequence, we fix $B=50$ and we track the range of the corresponding MSE values. With all being said, in the rest of this section, the performance of each method is evaluated with respect to (i) the MSE, (ii) the range of the SSE values obtained for each model, and (iii) the computational cost of the $B$ times (measured in seconds). In particular, in Tables  \ref{small_k3}--\ref{small_k10} the  differences of the minimum MSE values are illustrated, along with the ratios corresponding to the range of the MSE and the computational cost. Bold values indicate cases where \textsf{nnmf} performs better with respect to the minimum MSE values (positive differences), or the ratio of the Range and Time). Specifically, from Table  \ref{small_k3} we may see that, for $k=3$:
\begin{itemize}
\item The \textsf{nnmf} package is superior to the \textsf{RcppML} approach in the $84.6\%$ $(34.6\%)$ of the cases based on the MSE values (positive differences).
\item Based on the range of the MSE values during the $50$ iterations, \textsf{nnmf} is superior to the \textsf{RcppML} approach in the $92.3\%$ $(76.9\%)$ of the cases  (ratios higher than one).
\item The performance between the \textsf{RcppML} and \textsf{nnmf} packages is almost similar with respect to the computational cost. Strictly speaking the \textsf{RcppML} is a little faster. On the other hand, it can be seen that the \textsf{nnmf} is faster compared to the \textsf{NMF} in the $88.5\%$ of the cases.
\end{itemize}
For the same datasets, in Table   \ref{small_k10} similar results are presented for $k=10$. It can be concluded that:

\begin{itemize}
\item The \textsf{nnmf} package is superior to the \textsf{RcppML} package in the $76.9\%$ $(88.5\%)$ of the cases based on the MSE values (positive differences).
\item Based on the range of the MSE values during the $50$ iterations, \textsf{nnmf} is superior to\textsf{RcppML} in the $84.6\%$ $(80.8\%)$ of the cases (ratios higher than one).
\item The performance between \textsf{RcppML} and \textsf{nnmf} packages are almost similar with respect to the computational cost. Strictly speaking the \textsf{RcppML} is a little bit faster. On the other hand, it can be seen that the \textsf{nnmf} is faster than \textsf{NMF} in the $38.5\%$ of the cases but still the difference in computational time is not that large.
\end{itemize}

In summary, for the investigated cases presented in Table \ref{tab:datasets_info1}, the \textsf{nnmf} provides smaller MSE values with significant small MSEs' ranges in most of the cases compared to the \textsf{RcppML}. Strictly speaking, the \textsf{RcppML} is a little faster but the needed time does not significantly change. Furthermore, compared to \textsf{NMF}, even though the \textsf{nnmf} is superior in the  $34.6\%$ ($38.5\%$) when $k=3$ ($k=10$) of the cases, its performance is significantly better based on the range of the MSEs' and the computational time.

\begin{table}[htbp]
\centering
\caption{Datasets Information from UCI Machine Learning Repository and Kaggle database of small to moderate size.}
\label{tab:datasets_info1}
\begin{tabular}{lll}
\toprule
\textbf{Dataset ID} & \textbf{Title} & \textbf{Dimensions} \\ \midrule
1 & \href{https://doi.org/10.24432/C5NS3V}{Online News Popularity} & $39,644\times 52$ \\
2 & \href{https://doi.org/10.24432/C5004D}{Taiwanese Bankruptcy Prediction} & $6,819\times 95$ \\
3 & \href{https://doi.org/10.24432/C5XS4Q}{Sales Transactions Weekly} & $811\times 104$ \\
4 & \href{https://doi.org/10.24432/C53P47}{Superconductivty Data} & $21,263 \times 81$ \\
5 & \href{https://doi.org/10.1097/md.0000000000037258}{Gallstone} & $319 \times 34$ \\
6 & \href{https://doi.org/10.24432/C5GG7Q}{Large--scale Wave Energy Farm (1)} & $36,043\times 149$ \\
7 & \href{https://doi.org/10.24432/C5989V}{Hepatitis C Virus (HCV) for Egyptian patients} & $1,377 \times 29$ \\
8 & \href{https://doi.org/10.24432/C55W3J}{Paddy Dataset} & $2,789 \times  37$ \\
9 & \href{https://doi.org/10.24432/C51S51}{APS Failure at Scania Trucks} & $592 \times 168$ \\
10 & \href{https://doi.org/10.24432/C56W5M}{PIRvision FoG presence detection} & $7,650 \times 57$ \\
11 & \href{https://doi.org/10.24432/C5KP6B}{Codon usage} & $1,3027 \times  65$ \\
12 & \href{https://doi.org/10.24432/C5GG7Q}{Large--scale Wave Energy Farm (2)} & $7,277\times 302$ \\
13 & \href{https://doi.org/10.24432/C5GG7Q}{Large--scale Wave Energy Farm (3)} & $17,964\times 149$ \\
14 & \href{https://doi.org/10.24432/C5GG7Q}{Large--scale Wave Energy Farm (4)} & $2,318 \times 302$ \\ 
15 & \href{https://doi.org/10.24432/C50S4B}{Dry Bean} & $13,611\times 16$ \\
16 & \href{https://doi.org/10.24432/C5RC9X}{First-order theorem proving} & $6,118\times 53$ \\
17 & \href{https://doi.org/10.24432/C5XP52}{Gender Gap in Spanish WP} & $4,746\times 20$ \\
18 & \href{https://doi.org/10.24432/C5161N}{SkillCraft1 Master Table Dataset} & $3,395\times 17$ \\
19 & \href{https://doi.org/10.24432/C5004D}{Taiwanese Bankruptcy Prediction} & $6,819\times 96$ \\
20 & \href{https://doi.org/10.24432/C5S02S}{Turkiye Student Evaluation} & $5,820\times 33$ \\
21 & \href{https://doi.org/10.24432/C5PC8X}{Communities and Crime} & $2,215\times 104$ \\
22 & \href{https://doi.org/10.24432/C5K31K}{Condition Based Maintenance of Naval Propulsion Plants} & $11,934\times 18$ \\
23 & \href{https://doi.org/10.24432/C5ZS3N}{Parkinsons Telemonitoring} & $5,875\times 21$ \\
24 & \href{https://doi.org/10.24432/C5RS3S}{SML2010 (1)} & $2,764\times 21$ \\
25 & \href{https://doi.org/10.24432/C5RS3S}{SML2010 (2)} & $1,373\times 21$ \\
26 & \href{https://doi.org/10.24432/C53G6X}{Spambase} & $4,601\times 58$ \\ \bottomrule
\end{tabular}
\end{table}

\begin{table}[H]
\caption{Performance comparisons for $k=3$ based on the datasets presented in Table \ref{tab:datasets_info1}. For the MSE we compute the difference between \textsf{RcppML} and \textsf{NMF} with \textsf{nnmf}. For the range of the MSE and the running time, we compute the ratio of \textsf{RcppML} and \textsf{NMF} to \textsf{nnmf}.}
\centering
\resizebox{\textwidth}{!}{%
\begin{tabular}{c|ll|ll|rr}
  \hline
\textbf{Dataset} & \multicolumn{2}{c|}{\textbf{MSE}} & \multicolumn{2}{c|}{\textbf{Range}} & \multicolumn{2}{c}{\textbf{Running time}} \\
   & \textsf{RcppML}--\textsf{nnmf} & \textsf{NMF}--\textsf{nnmf} & \textsf{RcppML}/\textsf{nnmf} & \textsf{NMF}/\textsf{nnmf} & \textsf{RcppML}/\textsf{nnmf} & \textsf{NMF}/\textsf{nnmf} \\ \midrule
1 & $\mathbf{1.244\times10^{12}}$ & $-1.131\times10^{11}$ & $\mathbf{1.338}$ & $\mathbf{1.001}$ & $0.219$ & $\mathbf{58.009}$ \\ 
2 & $\mathbf{1.285\times10^{20}}$ & $-3.684\times10^{19}$ & $0.080$ & $0.477$ & $0.271$ & $\mathbf{89.364}$ \\ 
3 & $\mathbf{1.863\times10^{2}}$ & $-4.402\times10^{2}$ & $\mathbf{3.769\times10^{11}}$ & $\mathbf{5.030\times10^{10}}$ & $0.080$ & $\mathbf{17.315}$ \\ 
4 & $\mathbf{1.654\times10^{9}}$ & $-6.032\times10^{6}$ & $\mathbf{1.370\times10^{13}}$ & $\mathbf{1.860\times10^{10}}$ & $0.212$ & $\mathbf{66.897}$ \\ 
5 & $\mathbf{5.262\times10^{3}}$ & $-4.652\times10^{2}$ & $\mathbf{3.151}$ & $2.912\times10^{-3}$ & $0.193$ & $\mathbf{23.363}$ \\ 
6 & $-3.649\times10^{9}$ & $\mathbf{2.381\times10^{13}}$ & $\mathbf{4.259\times10^{8}}$ & $\mathbf{4.255\times10^{10}}$ & $0.373$ & $\mathbf{318.099}$ \\ 
7 & $\mathbf{5.975\times10^{11}}$ & $-4.072\times10^{11}$ & $\mathbf{3.906}$ & $0.652$ & $0.046$ & $\mathbf{3.956}$ \\ 
8 & $\mathbf{3.164\times10^{5}}$ & $\mathbf{2.132\times10^{7}}$ & $\mathbf{8.891\times10^{1}}$ & $\mathbf{3.992\times10^{3}}$ & $0.044$ & $\mathbf{8.409}$ \\ 
9 & $\mathbf{5.099\times10^{15}}$ & $-1.337\times10^{14}$ & $\mathbf{1.011}$ & $\mathbf{1.009}$ & $0.015$ & $\mathbf{9.963}$ \\ 
10 & $-2.259\times10^{10}$ & $-9.289\times10^{7}$ & $\mathbf{4.483\times10^{8}}$ & $\mathbf{6.899\times10^{11}}$ & $0.001$ & $\mathbf{1.059}$ \\ 
11 & $-2.626\times10^{10}$ & $-1.419\times10^{8}$ & $\mathbf{4.920\times10^{8}}$ & $\mathbf{6.900\times10^{11}}$ & $0.001$ & $\mathbf{1.063}$ \\ 
12 & $\mathbf{1.438\times10^{9}}$ & $\mathbf{1.540\times10^{13}}$ & $\mathbf{4.950}$ & $\mathbf{6.129\times10^{2}}$ & $0.239$ & $\mathbf{216.896}$ \\ 
13 & $-4.114\times10^{9}$ & $\mathbf{2.894\times10^{12}}$ & $\mathbf{6.184\times10^{1}}$ & $\mathbf{6.334\times10^{3}}$ & $0.262$ & $\mathbf{167.890}$ \\ 
14 & $\mathbf{2.070\times10^{10}}$ & $\mathbf{5.165\times10^{12}}$ & $\mathbf{1.867\times10^{1}}$ & $\mathbf{6.846\times10^{3}}$ & $0.319$ & $\mathbf{199.958}$ \\ 
15 & $\mathbf{8.878\times10^{2}}$ & $\mathbf{7.177\times10^{4}}$ & $\mathbf{2.233\times10^{4}}$ & $\mathbf{7.437\times10^{5}}$ & $0.049$ & $\mathbf{5.299}$ \\
16 & $\mathbf{1.488\times10^{1}}$ & $-3.958\times10^{-2}$ & $\mathbf{4.564\times10^{1}}$ & $1.754\times10^{-2}$ & $0.020$ & $\mathbf{11.150}$ \\
17 & $\mathbf{5.340\times10^{3}}$ & $-1.765\times10^{3}$ & $\mathbf{3.469\times10^{13}}$ & $\mathbf{6.859\times10^{9}}$ & $0.022$ & $\mathbf{11.700}$ \\
18 & $\mathbf{2.834\times10^{2}}$ & $\mathbf{1.744\times10^{3}}$ & $\mathbf{3.772\times10^{10}}$ & $\mathbf{2.326\times10^{11}}$ & $0.010$ & $0.511$ \\
19 & $\mathbf{1.740\times10^{16}}$ & $-5.403\times10^{15}$ & $0.482$ & $0.477$ & $0.306$ & $\mathbf{106.700}$ \\
20 & $\mathbf{6.538\times10^{-2}}$ & $-9.580\times10^{-3}$ & $\mathbf{1.453\times10^{2}}$ & $1.807\times10^{-1}$ & $0.349$ & $\mathbf{31.560}$ \\
21 & $\mathbf{4.069\times10^{6}}$ & $-8.542\times10^{5}$ & $\mathbf{2.104\times10^{12}}$ & $\mathbf{2.638\times10^{12}}$ & $0.055$ & $\mathbf{20.670}$ \\
22 & $\mathbf{2.439\times10^{4}}$ & $\mathbf{1.095\times10^{4}}$ & $\mathbf{7.619\times10^{4}}$ & $\mathbf{4.859\times10^{2}}$ & $0.059$ & $\mathbf{6.415}$ \\
23 & $\mathbf{1.065\times10^{0}}$ & $-4.351\times10^{-2}$ & $\mathbf{9.392\times10^{1}}$ & $\mathbf{1.802\times10^{0}}$ & $0.065$ & $\mathbf{6.501}$ \\
24 & $\mathbf{6.989\times10^{5}}$ & $-3.635\times10^{2}$ & $\mathbf{2.072\times10^{12}}$ & $\mathbf{5.783\times10^{8}}$ & $0.001$ & $0.101$ \\
25 & $\mathbf{4.573\times10^{5}}$ & $-2.791\times10^{2}$ & $\mathbf{1.745\times10^{5}}$ & $\mathbf{7.410\times10^{1}}$ & $0.001$ & $0.185$ \\
26 & $\mathbf{1.313\times10^{0}}$ & $\mathbf{8.478\times10^{-2}}$ & $\mathbf{3.667\times10^{12}}$ & $\mathbf{3.657\times10^{12}}$ & $0.193$ & $\mathbf{52.360}$ \\
   \hline
\textbf{\% \textsf{nnmf} outperforms} & \textbf{84.6\%} & \textbf{34.6\%} & \textbf{92.3\%} & \textbf{76.9\%} & \textbf{0.0\%} & \textbf{88.5\%} \\
   \hline
\end{tabular}}
   \label{small_k3}
\end{table}

\begin{table}[H]
\caption{Performance comparisons for $k=10$ based on the datasets presented in Table \ref{tab:datasets_info1}. For the MSE we compute the difference between \textsf{RcppML} and \textsf{NMF} with \textsf{nnmf}. For the range of the MSE and the running time, we compute the ratio of \textsf{RcppML} and \textsf{NMF} to \textsf{nnmf}.
}
\centering
\resizebox{\textwidth}{!}{%
\begin{tabular}{c|ll|ll|rr}
  \hline
\textbf{Dataset} & \multicolumn{2}{c|}{\textbf{MSE}} & \multicolumn{2}{c|}{\textbf{Range}} & \multicolumn{2}{c}{\textbf{Running time}} \\
   & \textsf{RcppML}--\textsf{nnmf} & \textsf{NMF}--\textsf{nnmf} & \textsf{RcppML}/\textsf{nnmf} & \textsf{NMF}/\textsf{nnmf} & \textsf{RcppML}/\textsf{nnmf} & \textsf{NMF}/\textsf{nnmf} \\ \midrule
1 & $\mathbf{8.424\times10^{5}}$ & $\mathbf{6.879\times10^{7}}$ & $\mathbf{34.237}$ & $\mathbf{423.933}$ & $0.076$ & $0.254$ \\ 
2 & $\mathbf{2.344\times10^{16}}$ & $-1.078\times10^{14}$ & $\mathbf{1.009}$ & $0.952$ & $0.004$ & $\mathbf{1.543}$ \\ 
3 & $\mathbf{0.797}$ & $-0.655$ & $\mathbf{1.930}$ & $0.695$ & $0.031$ & $\mathbf{11.415}$ \\ 
4 & $\mathbf{2.618\times10^{4}}$ & $\mathbf{3.968\times10^{3}}$ & $\mathbf{9.903}$ & $\mathbf{5.693}$ & $0.014$ & $\mathbf{4.037}$ \\ 
5 & $\mathbf{46.450}$ & $\mathbf{0.108}$ & $\mathbf{17.889}$ & $\mathbf{4.752}$ & $0.006$ & $\mathbf{1.096}$ \\ 
6 & $\mathbf{1.089\times10^{7}}$ & $\mathbf{1.183\times10^{9}}$ & $\mathbf{4.769}$ & $\mathbf{173.399}$ & $0.182$ & $\mathbf{107.130}$ \\ 
7 & $-6.579\times10^{6}$ & $\mathbf{1.487\times10^{7}}$ & $0.004$ & $0.137$ & $0.132$ & $0.028$ \\ 
8 & $\mathbf{2.566}$ & $\mathbf{1.248\times10^{3}}$ & $\mathbf{6.906}$ & $\mathbf{386.883}$ & $0.015$ & $0.124$ \\ 
9 & $\mathbf{3.439\times10^{13}}$ & $\mathbf{3.890\times10^{14}}$ & $\mathbf{3.177}$ & $\mathbf{25.778}$ & $0.021$ & $0.842$ \\ 
10 & $-5.496\times10^{5}$ & $\mathbf{8.029\times10^{6}}$ & $\mathbf{2.927}$ & $\mathbf{3.502}$ & $0.003$ & $0.122$ \\ 
11 & $-2.041\times10^{5}$ & $\mathbf{8.664\times10^{6}}$ & $\mathbf{2.219}$ & $\mathbf{2.577}$ & $0.006$ & $0.137$ \\ 
12 & $\mathbf{3.805\times10^{7}}$ & $\mathbf{4.876\times10^{9}}$ & $\mathbf{1.791}$ & $\mathbf{42.410}$ & $0.276$ & $\mathbf{132.021}$ \\ 
13 & $\mathbf{3.245\times10^{7}}$ & $\mathbf{2.866\times10^{9}}$ & $\mathbf{9.541}$ & $\mathbf{270.629}$ & $0.196$ & $\mathbf{64.983}$ \\ 
14 & $\mathbf{8.359\times10^{6}}$ & $\mathbf{6.638\times10^{8}}$ & $\mathbf{8.121}$ & $\mathbf{651.432}$ & $0.210$ & $\mathbf{100.266}$ \\ 
15 & $\mathbf{2.970}$ & $\mathbf{824.933}$ & $\mathbf{3.486}$ & $\mathbf{498.886}$ & $0.011$ & $0.236$ \\ 
16 & $\mathbf{6.934}$ & $\mathbf{19.473}$ & $\mathbf{21.148}$ & $\mathbf{7.428}$ & $5.050\times10^{-4}$ & $0.114$ \\ 
17 & $\mathbf{8447.005}$ & $\mathbf{5073.001}$ & $\mathbf{4.006\times10^{6}}$ & $\mathbf{3.610\times10^{6}}$ & $9.745\times10^{-4}$ & $0.266$ \\ 
18 & $\mathbf{0.479}$ & $\mathbf{176.526}$ & $\mathbf{1.288}$ & $\mathbf{14.146}$ & $0.037$ & $0.273$ \\ 
19 & $\mathbf{3.371\times10^{16}}$ & $-7.874\times10^{13}$ & $\mathbf{1.067}$ & $0.962$ & $0.007$ & $\mathbf{2.684}$ \\ 
20 & $\mathbf{0.110}$ & $\mathbf{2.880\times10^{-3}}$ & $\mathbf{7.410}$ & $\mathbf{5.314}$ & $0.037$ & $\mathbf{2.107}$ \\ 
21 & $\mathbf{2.209\times10^{7}}$ & $\mathbf{3.272\times10^{6}}$ & $\mathbf{4.636}$ & $\mathbf{1.450}$ & $0.004$ & $0.589$ \\ 
22 & $\mathbf{17.503}$ & $\mathbf{2.090\times10^{3}}$ & $\mathbf{2.492}$ & $\mathbf{5.312}$ & $0.014$ & $0.310$ \\ 
23 & $-0.025$ & $\mathbf{0.165}$ & $0.010$ & $0.018$ & $0.311$ & $0.134$ \\ 
24 & $-67.671$ & $\mathbf{6.472\times10^{4}}$ & $0.454$ & $\mathbf{1.697}$ & $0.048$ & $0.023$ \\ 
25 & $-181.948$ & $\mathbf{5.156\times10^{4}}$ & $0.648$ & $\mathbf{3.204}$ & $0.387$ & $0.132$ \\ 
26 & $\mathbf{1.085}$ & $\mathbf{11.094}$ & $\mathbf{1.696\times10^{10}}$ & $\mathbf{2.099\times10^{10}}$ & $0.004$ & $0.199$ \\ 
   \hline
\textbf{\% nnmf better} & \textbf{76.9\%} & \textbf{88.5\%} & \textbf{84.6\%} & \textbf{80.8\%} & \textbf{0.0\%} & \textbf{38.5\%} \\
   \hline
\end{tabular}}
   \label{small_k10}
\end{table}

\section{Conclusions}
We focused on NMF using the \textit{R} package \textsf{nnmf} which we compared with the established \textsf{NMF} package and the \textsf{RcppML} package. The common ground of all packages is their \textit{C++} implementation of the NMF. The empirical comparison took place using real data to see the actual performance of the three packages, with regards three pillars, the reconstruction error, the range of the solutions produced upon running NMF multiple times and the running time. The rank of the lower representation was set equal to $k=3$ and $k=10$. 

When $k=3$, \textsf{NMF} yielded the lowest SSE, followed by \textsf{nnmf} and then \textsf{RcppML}. The range of the \textsf{nnmf} was the smallest among its competitors, for the majority of the datasets. \textsf{RcppML} contains the fastest implementation, followed \textsf{nnmf} and then \textsf{NMF}. When $k=10$, \textsf{nnmf} produced the lowest SSE in most cases, and the range was again the smallest. \textsf{RcppML} remains the fastest option, but the running time of \textsf{nnmf} increased compared to \textsf{NMF}. Overall, \textsf{RcppML} is the universal winner in terms of computational cost and \textsf{nnmf} is the winner in terms of variability of the solutions produced. In terms of reconstruction error, \textsf{NMF} is preferred for low ranks, whereas for higher ranks \textsf{nnmf} is to be preferred. 

\bibliographystyle{apalike}
\bibliography{references}

\clearpage
\section{Appendix}

\begin{table}[H]
\caption{Raw values for $k=3$ based on the datasets presented in Table \ref{tab:datasets_info1}.}
\centering
\resizebox{\textwidth}{!}{%
\begin{tabular}{c|lll|lll|rrr}
  \toprule
\textbf{Dataset} & \multicolumn{3}{c|}{\textbf{MSE}} & \multicolumn{3}{c|}{\textbf{Range}} & \multicolumn{3}{c}{\textbf{Runing time}} \\
 & \textsf{nnmf} & \textsf{RcppML} & \textsf{NMF} & \textsf{nnmf} & \textsf{RcppML} & \textsf{NMF} & \textsf{nnmf} & \textsf{RcppML} & \textsf{NMF} \\  \midrule
1 & $1.122\times10^{14}$ & $1.134\times10^{14}$ & $1.121\times10^{14}$ & $1.753\times10^{13}$ & $2.345\times10^{13}$ & $1.754\times10^{13}$ & $47.530$ & $10.396$ & $2757.171$ \\ 
2 & $2.532\times10^{23}$ & $2.534\times10^{23}$ & $2.532\times10^{23}$ & $2.547\times10^{22}$ & $2.046\times10^{21}$ & $1.216\times10^{22}$ & $8.143$ & $2.209$ & $727.712$ \\ 
3 & $3.946\times10^{5}$ & $3.948\times10^{5}$ & $3.941\times10^{5}$ & $7.451\times10^{-9}$ & $2.808\times10^{3}$ & $3.748\times10^{2}$ & $4.710$ & $0.375$ & $81.547$ \\ 
4 & $6.208\times10^{10}$ & $6.373\times10^{10}$ & $6.207\times10^{10}$ & $1.953\times10^{-3}$ & $2.675\times10^{10}$ & $3.633\times10^{7}$ & $35.350$ & $7.482$ & $2364.830$ \\ 
5 & $2.141\times10^{6}$ & $2.147\times10^{6}$ & $2.141\times10^{6}$ & $3.008\times10^{3}$ & $9.477\times10^{3}$ & $8.758$ & $0.711$ & $0.137$ & $16.604$ \\ 
6 & $1.560\times10^{14}$ & $1.560\times10^{14}$ & $1.798\times10^{14}$ & $5.760\times10^{2}$ & $2.453\times10^{11}$ & $2.451\times10^{13}$ & $32.327$ & $12.067$ & $10283.230$ \\ 
7 & $2.833\times10^{14}$ & $2.839\times10^{14}$ & $2.829\times10^{14}$ & $1.186\times10^{13}$ & $4.631\times10^{13}$ & $7.730\times10^{12}$ & $13.066$ & $0.599$ & $51.691$ \\ 
8 & $3.513\times10^{6}$ & $3.829\times10^{6}$ & $2.483\times10^{7}$ & $2.629\times10^{4}$ & $2.338\times10^{6}$ & $1.050\times10^{8}$ & $12.779$ & $0.558$ & $107.457$ \\ 
9 & $1.815\times10^{18}$ & $1.820\times10^{18}$ & $1.815\times10^{18}$ & $2.207\times10^{18}$ & $2.232\times10^{18}$ & $2.226\times10^{18}$ & $9.016$ & $0.138$ & $89.830$ \\ 
10 & $1.174\times10^{11}$ & $9.485\times10^{10}$ & $1.173\times10^{11}$ & $64.000$ & $2.869\times10^{10}$ & $4.416\times10^{13}$ & $411.608$ & $0.528$ & $435.738$ \\ 
11 & $1.174\times10^{11}$ & $9.117\times10^{10}$ & $1.173\times10^{11}$ & $64.000$ & $3.149\times10^{10}$ & $4.416\times10^{13}$ & $410.142$ & $0.469$ & $435.880$ \\ 
12 & $1.090\times10^{14}$ & $1.090\times10^{14}$ & $1.244\times10^{14}$ & $1.717\times10^{10}$ & $8.499\times10^{10}$ & $1.052\times10^{13}$ & $14.016$ & $3.355$ & $3040.084$ \\ 
 13& $1.570\times10^{13}$ & $1.569\times10^{13}$ & $1.859\times10^{13}$ & $5.816\times10^{8}$ & $3.596\times10^{10}$ & $3.684\times10^{12}$ & $3.978$ & $1.044$ & $667.933$ \\ 
14 & $2.537\times10^{13}$ & $2.540\times10^{13}$ & $3.054\times10^{13}$ & $2.153\times10^{9}$ & $4.020\times10^{10}$ & $1.474\times10^{13}$ & $15.787$ & $5.032$ & $3156.761$ \\
  115 & $7.676\times10^{2}$ & $1.655\times10^{3}$ & $7.254\times10^{4}$ & $7.517\times10^{-2}$ & $1.678\times10^{3}$ & $5.590\times10^{4}$ & $69.147$ & $3.356$ & $366.409$ \\ 
16 & $9.043\times10^{2}$ & $9.192\times10^{2}$ & $9.042\times10^{2}$ & $1.548$ & $70.649$ & $2.716\times10^{-2}$ & $29.481$ & $0.584$ & $328.722$ \\ 
17 & $1.591\times10^{6}$ & $1.597\times10^{6}$ & $1.590\times10^{6}$ & $5.146\times10^{-8}$ & $1.785\times10^{6}$ & $3.529\times10^{2}$ & $38.336$ & $0.846$ & $448.637$ \\
18 & $1.841\times10^{2}$ & $4.675\times10^{2}$ & $1.928\times10^{3}$ & $2.247\times10^{-8}$ & $8.476\times10^{2}$ & $5.226\times10^{3}$ & $171.345$ & $1.656$ & $87.561$ \\ 
19 & $3.714\times10^{19}$ & $3.715\times10^{19}$ & $3.713\times10^{19}$ & $3.735\times10^{18}$ & $1.800\times10^{18}$ & $1.783\times10^{18}$ & $6.920$ & $2.120$ & $738.256$ \\ 
20 & $10.632$ & $10.698$ & $10.623$ & $5.005\times10^{-3}$ & $0.727$ & $9.045\times10^{-4}$ & $6.860$ & $2.394$ & $216.529$ \\ 
21 & $1.490\times10^{9}$ & $1.494\times10^{9}$ & $1.489\times10^{9}$ & $1.130\times10^{-4}$ & $2.377\times10^{8}$ & $2.982\times10^{8}$ & $10.383$ & $0.572$ & $214.631$ \\ 
22 & $1.947\times10^{3}$ & $2.634\times10^{4}$ & $1.290\times10^{4}$ & $4.992$ & $3.803\times10^{5}$ & $2.426\times10^{3}$ & $50.410$ & $2.989$ & $323.373$ \\ 
23 & $30.811$ & $31.876$ & $30.767$ & $5.327\times10^{-2}$ & $5.003$ & $9.599\times10^{-2}$ & $25.084$ & $1.634$ & $163.084$ \\ 
24 & $1.110\times10^{5}$ & $8.098\times10^{5}$ & $1.106\times10^{5}$ & $1.060\times10^{-6}$ & $2.197\times10^{6}$ & $6.129\times10^{2}$ & $754.757$ & $0.450$ & $76.484$ \\ 
25 & $1.012\times10^{5}$ & $5.586\times10^{5}$ & $1.010\times10^{5}$ & $6.042$ & $1.054\times10^{6}$ & $4.477\times10^{2}$ & $231.552$ & $0.207$ & $42.822$ \\ 
26 & $40.154$ & $41.468$ & $40.239$ & $2.073\times10^{-10}$ & $7.601\times10^{2}$ & $7.580\times10^{2}$ & $4.902$ & $0.945$ & $256.654$ \\ 
   \bottomrule
\end{tabular}}
\end{table}

\begin{table}[H]
\caption{Raw values for $k=10$ based on the datasets presented in Table \ref{tab:datasets_info1}.}
\centering
\resizebox{\textwidth}{!}{%
\begin{tabular}{c|lll|lll|rrr}
  \toprule
\textbf{Dataset} & \multicolumn{3}{c|}{\textbf{MSE}} & \multicolumn{3}{c|} 
{\textbf{Range}} & \multicolumn{3}{c}{\textbf{Runing time}} \\ 
 & \textsf{nnmf} & \textsf{RcppML} & \textsf{NMF} & \textsf{nnmf} & \textsf{RcppML} & \textsf{NMF} & \textsf{nnmf} & \textsf{RcppML} & \textsf{NMF} \\ 
  \midrule 
1 & $3.019\times10^{5}$ & $1.144\times10^{6}$ & $6.910\times10^{7}$ & $9.928\times10^{5}$ & $3.399\times10^{7}$ & $4.209\times10^{8}$ & $25906.761$ & $1973.932$ & $6569.593$ \\ 
2 & $8.065\times10^{17}$ & $8.299\times10^{17}$ & $8.064\times10^{17}$ & $6.943\times10^{17}$ & $7.009\times10^{17}$ & $6.607\times10^{17}$ & $1187.118$ & $4.587$ & $1831.137$ \\ 
3 & $3.664\times10^{2}$ & $3.672\times10^{2}$ & $3.657\times10^{2}$ & $1.642$ & $3.168$ & $1.141$ & $19.097$ & $0.597$ & $218.002$ \\ 
4 & $6.335\times10^{4}$ & $8.954\times10^{4}$ & $6.732\times10^{4}$ & $1.422\times10^{4}$ & $1.408\times10^{5}$ & $8.095\times10^{4}$ & $1344.986$ & $18.805$ & $5430.284$ \\ 
5 & $4.198\times10^{2}$ & $4.663\times10^{2}$ & $4.199\times10^{2}$ & $4.489$ & $80.304$ & $21.331$ & $39.451$ & $0.238$ & $43.220$ \\ 
6 & $2.257\times10^{9}$ & $2.268\times10^{9}$ & $3.439\times10^{9}$ & $5.328\times10^{6}$ & $2.541\times10^{7}$ & $9.238\times10^{8}$ & $179.006$ & $32.514$ & $19176.918$ \\ 
7 & $6.584\times10^{6}$ & $5.182\times10^{3}$ & $2.146\times10^{7}$ & $5.849\times10^{9}$ & $2.319\times10^{7}$ & $7.996\times10^{8}$ & $4976.219$ & $659.295$ & $137.194$ \\ 
8 & $4.561\times10^{-3}$ & $2.571$ & $1.248\times10^{3}$ & $18.322$ & $1.265\times10^{2}$ & $7.088\times10^{3}$ & $2409.084$ & $36.025$ & $299.885$ \\ 
9 & $1.756\times10^{14}$ & $2.100\times10^{14}$ & $5.646\times10^{14}$ & $9.418\times10^{13}$ & $2.993\times10^{14}$ & $2.428\times10^{15}$ & $289.355$ & $6.150$ & $243.517$ \\ 
10 & $2.641\times10^{6}$ & $2.092\times10^{6}$ & $1.067\times10^{7}$ & $1.328\times10^{6}$ & $3.888\times10^{6}$ & $4.652\times10^{6}$ & $9781.088$ & $25.986$ & $1196.010$ \\ 
11 & $2.220\times10^{6}$ & $2.016\times10^{6}$ & $1.088\times10^{7}$ & $1.750\times10^{6}$ & $3.882\times10^{6}$ & $4.508\times10^{6}$ & $8751.977$ & $54.407$ & $1195.209$ \\ 
12 & $7.882\times10^{9}$ & $7.920\times10^{9}$ & $1.276\times10^{10}$ & $5.079\times10^{7}$ & $9.098\times10^{7}$ & $2.154\times10^{9}$ & $48.221$ & $13.317$ & $6366.139$ \\ 
13 & $2.942\times10^{9}$ & $2.975\times10^{9}$ & $5.809\times10^{9}$ & $6.899\times10^{6}$ & $6.582\times10^{7}$ & $1.867\times10^{9}$ & $25.797$ & $5.066$ & $1676.386$ \\ 
14 & $3.775\times10^{8}$ & $3.859\times10^{8}$ & $1.041\times10^{9}$ & $1.097\times10^{6}$ & $8.909\times10^{6}$ & $7.147\times10^{8}$ & $74.668$ & $15.666$ & $7486.681$ \\
15 & $3.442\times10^{-3}$ & $2.974$ & $8.249\times10^{2}$ & $1.459\times10^{2}$ & $5.084\times10^{2}$ & $7.276\times10^{4}$ & $4399.345$ & $47.933$ & $1038.235$ \\ 
16 & $5.337$ & $12.272$ & $24.810$ & $6.629$ & $1.402\times10^{2}$ & $49.241$ & $7868.949$ & $3.974$ & $900.543$ \\ 
17 & $1.741\times10^{4}$ & $2.586\times10^{4}$ & $2.248\times10^{4}$ & $2.094\times10^{-2}$ & $8.387\times10^{4}$ & $7.557\times10^{4}$ & $5425.087$ & $5.287$ & $1441.939$ \\
18 & $1.858$ & $2.337$ & $1.784\times10^{2}$ & $39.086$ & $50.325$ & $5.529\times10^{2}$ & $904.322$ & $33.502$ & $246.886$ \\ 
19 & $8.064\times10^{17}$ & $8.402\times10^{17}$ & $8.064\times10^{17}$ & $6.719\times10^{17}$ & $7.167\times10^{17}$ & $6.466\times10^{17}$ & $697.096$ & $4.773$ & $1871.073$ \\ 
20 & $4.185$ & $4.296$ & $4.188$ & $3.696\times10^{-2}$ & $0.274$ & $0.196$ & $288.667$ & $10.543$ & $608.099$ \\ 
21 & $7.409\times10^{7}$ & $9.618\times10^{7}$ & $7.736\times10^{7}$ & $6.749\times10^{7}$ & $3.129\times10^{8}$ & $9.783\times10^{7}$ & $1001.671$ & $4.109$ & $590.476$ \\ 
22 & $0.289$ & $17.793$ & $2.090\times10^{3}$ & $1.997\times10^{3}$ & $4.976\times10^{3}$ & $1.061\times10^{4}$ & $3001.053$ & $41.776$ & $931.156$ \\ 
23 & $2.940\times10^{-2}$ & $4.498\times10^{-3}$ & $0.195$ & $3.577$ & $3.646\times10^{-2}$ & $6.562\times10^{-2}$ & $3505.267$ & $1089.503$ & $469.688$ \\ 
24 & $3.763\times10^{2}$ & $3.086\times10^{2}$ & $6.509\times10^{4}$ & $2.561\times10^{4}$ & $1.162\times10^{4}$ & $4.347\times10^{4}$ & $9563.768$ & $461.924$ & $217.658$ \\ 
25 & $3.073\times10^{2}$ & $1.254\times10^{2}$ & $5.187\times10^{4}$ & $1.326\times10^{4}$ & $8.589\times10^{3}$ & $4.250\times10^{4}$ & $836.263$ & $323.516$ & $110.605$ \\ 
26 & $11.492$ & $12.578$ & $22.586$ & $8.291\times10^{-10}$ & $14.059$ & $17.401$ & $3531.084$ & $12.363$ & $703.005$ \\ 
   \bottomrule
\end{tabular}}

\end{table}

\end{document}